\documentclass[11pt,a4paper]{article}
\usepackage[hyperref]{acl2019}

\usepackage{amsmath,amsfonts,bm}

\def\eqref#1{equation~\ref{#1}}

\def\1{\bm{1}}

\DeclareMathAlphabet{\mathsfit}{\encodingdefault}{\sfdefault}{m}{sl}
\SetMathAlphabet{\mathsfit}{bold}{\encodingdefault}{\sfdefault}{bx}{n}

\usepackage{times}
\usepackage{latexsym}
\usepackage{booktabs}
\usepackage{hyperref}
\usepackage{url}
\usepackage{graphicx}
\usepackage{xcolor}
\usepackage{multirow}
\usepackage[normalem]{ulem}
\usepackage{wrapfig}
\usepackage{color, colortbl}

\aclfinalcopy \def\aclpaperid{1913} 

\definecolor{Gray}{gray}{0.93}
\definecolor{LightGray}{gray}{1.0}
\newcommand\datasetname{\textsc{EmpatheticDialogues}}
\newcommand\datasetabbrev{ED}
\newcommand\basename{Fine-Tuned}
\newcommand\coauth{$^\star$}
\title{Towards Empathetic Open-domain Conversation Models: a New Benchmark and Dataset}

\author{Hannah Rashkin$^1$\coauth{}, Eric Michael Smith$^2$, Margaret Li$^2$, Y-Lan Boureau$^2$\\
$^1$ Paul G. Allen School of Computer Science \& Engineering, University of Washington\\
$^2$ Facebook AI Research\\
\texttt{hrashkin@cs.washington.edu}, \texttt{\{ems,margaretli,ylan\}@fb.com} \\
}

\date{}

\begin{document}

\maketitle

\begin{abstract}
One challenge for dialogue agents is recognizing feelings in the conversation partner and replying accordingly, a key communicative skill. While it is straightforward for humans to recognize and acknowledge others' feelings in a conversation, this is a significant challenge for AI systems due to the paucity of suitable publicly-available datasets for training and evaluation. This work proposes a new benchmark for empathetic dialogue generation and \datasetname{}, a novel dataset  
of 25k conversations grounded in emotional situations.  
Our experiments indicate that dialogue models that use our dataset
are perceived to be more empathetic by human evaluators, 
compared to models
merely trained on large-scale Internet conversation data. We
also present empirical comparisons of dialogue model adaptations for empathetic responding, leveraging existing models or datasets without requiring lengthy re-training of the
full model. 
\let\thefootnote\relax\footnotetext{\coauth{}This work was done while first author was intern at Facebook AI Research (FAIR).}
\end{abstract}

\section{Introduction}

\begin{figure}
    \centering
    \includegraphics[trim={9.cm 12.75cm 13.cm 7cm},clip,width=.45\textwidth]{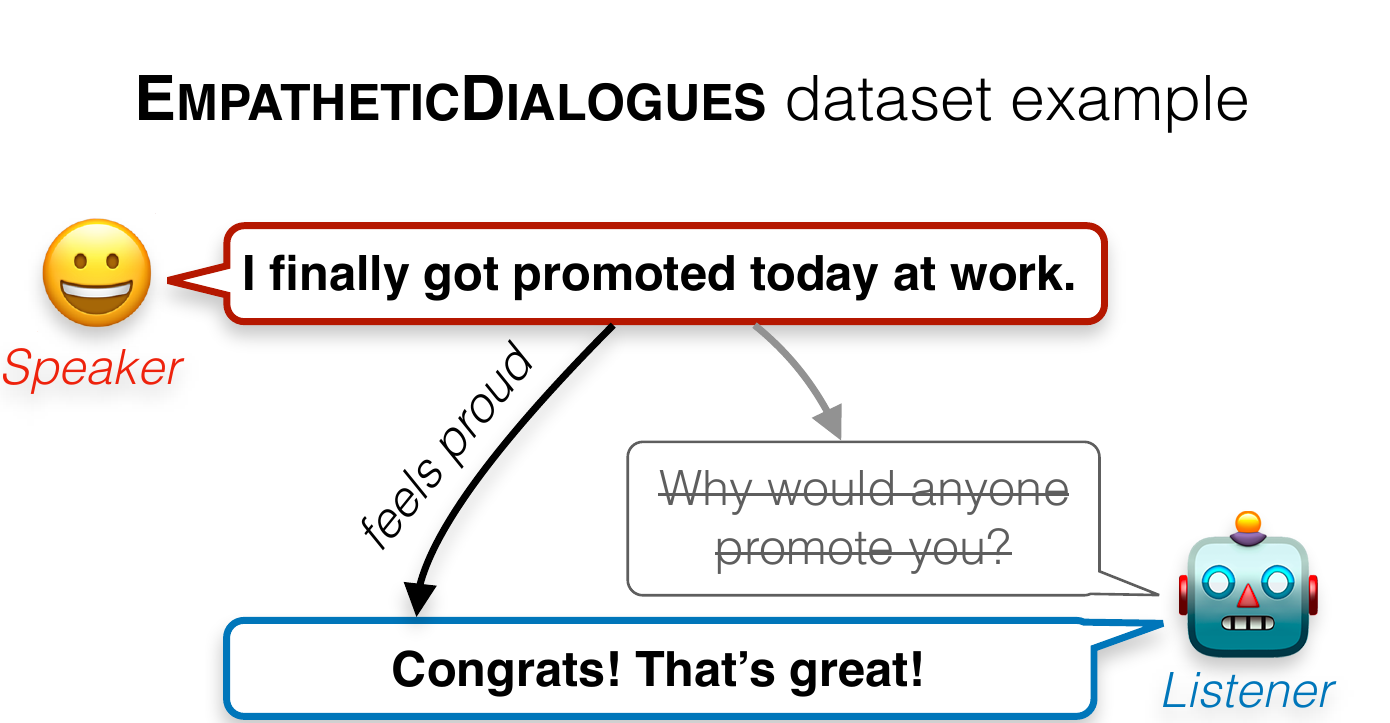}
    \caption{Example where acknowledging an inferred feeling is appropriate}
    \label{fig:convo}
\end{figure}

A desirable trait in a human-facing dialogue agent is to appropriately respond to a conversation partner that is describing personal experiences, by understanding and acknowledging any implied feelings --- a skill we refer to as empathetic responding.
For instance, while the crossed-out response in  Figure~\ref{fig:convo} is topically relevant, ``Congrats! That's great!'' may be more satisfying because it acknowledges the underlying feelings of accomplishment in an empathetic way.
In this work, we investigate empathetic response generation from current dialogue systems, and propose experiments using a new 
resource, \datasetname{}, as a benchmark to evaluate this skill set.
\begin{figure*}
\small
\begin{center}
\begin{tabular}{p{3.0in}p{3.1in}}
\hspace*{-2.mm}\fbox{\begin{tabular}[t]{p{2.75in}}\textbf{Label: Afraid}\\ \textbf{Situation:} Speaker felt this when...\\``I've been hearing noises around the house at night'' \\ \textbf{Conversation:}\\ \vspace*{-2.5mm}{\color{red} Speaker}: I've been hearing some strange noises around the house at night.\\\vspace*{-2.5mm} {\color{blue} Listener}: oh no! That's scary! What do you think it is?\\\vspace*{-2.5mm}{\color{red} Speaker}: I don't know, that's what's making me anxious.\\ \vspace*{-2.5mm}{\color{blue} Listener}: I'm sorry to hear that. I wish I could help you figure it out\vspace*{.6mm}\end{tabular}} &  \fbox{\begin{tabular}[t]{p{2.75in}}\textbf{Label: Proud}\tabularnewline \textbf{Situation:} Speaker felt this when...\tabularnewline``I finally got that promotion at work! I have tried so hard for so long to get it!'' \tabularnewline \textbf{Conversation:}\tabularnewline\vspace*{-2.5mm} {\color{red} Speaker}: I finally got promoted today at work!\tabularnewline \vspace*{-2.5mm}{\color{blue} Listener}: Congrats! That's great!\\\vspace*{-2.5mm} {\color{red} Speaker}: Thank you! I've been trying to get it for a while now!\tabularnewline\vspace*{-2.5mm} {\color{blue} Listener}: That is quite an accomplishment and you should be proud!\end{tabular}}
\end{tabular}
\caption{Two examples from \datasetname{} training set.  The first worker (the speaker) is given an emotion label and writes their own description of a situation when they've felt that way. Then, the speaker tells their story in a conversation with a second worker (the listener).}
\label{dataex}
\vspace*{-5mm}
\end{center}
\end{figure*}

Empathetic responding is clearly relevant to dialogue systems that are geared towards general conversation or chit-chat. Indeed, ordinary communication is frequently prompted by people sharing their feelings or circumstances. 
But researchers analyzing goal-directed conversations have also observed the frequent intrusion of ordinary conversation in those interactions as well, either as a ``warm-up'' introduction 
or as a detour 
\citep{levinson2000study,heritage2005conversation}. Engaging in social talk, reacting to emotional cues and displaying a caring attitude have, in fact, been associated with better task outcomes in many domains \citep{wentzel1997student,levinson2000study,bickmore2001relational,kim2004effects,fraser2018spoken}. While many of those studies deal with human-human interactions, humans have been shown to often interact with machines in a natural and social way \citep{reeves1996media,lee2010receptionist}, so it is reasonable to expect that dialogue agents would also benefit from empathetic responding.

Most recent powerful language architectures are trained on vast amounts of barely curated text scrapes, social media conversations, or independent books \citep{ritter2010unsupervised,personachat,mazare2018training,bertppr,liu2019multi,radford2019language}.
It might be the case that models trained on this type of data could exhibit some of the aggressive and callous responses that have been observed in spontaneous internet conversations \citep{anderson2015ask}. Unfortunately, while chitchat dialogue benchmarks have been proposed (e.g., \citealp{dinan2019second}),
to the best of our knowledge there are currently no benchmarks gauging whether dialogue agents can converse with empathy.

This work aims to facilitate evaluating models' ability to produce empathetic responses. We introduce a new task for dialogue systems to respond to people discussing situations that cover a wide range of emotions, and \datasetname{} (ED), a novel dataset with about 25k personal dialogues. Each dialogue is grounded in a specific situation where a speaker was feeling a given emotion, with a listener responding (Figure~\ref{dataex}). The new resource consists of crowdsourced one-on-one conversations, and covers a large set of emotions in a balanced way. 
This dataset is larger and contains a more extensive set of emotions than many similar emotion prediction datasets from other text domains such as \citet{ISEAR}, \citet{SemEval2007}, \citet{Mohammad2018SemEval2018T1}, and \citet{gupta2017sentiment}. The dataset has been publicly released, with code to reproduce the main experimental results of this paper\footnote{https://github.com/facebookresearch/EmpatheticDialogues}.

Our experiments show that large-capacity conversation models trained on spontaneous internet conversation data are not rated as very empathetic. We propose two simple ways to leverage our dataset to improve those models: use utterances from our training data as candidate responses in a retrieval model at inference time, and fine-tune the model on our task.
Finally, we explore whether different ways of combining information from related tasks can lead to more empathetic responses.
The contributions of this work are thus: 1) we release a novel empathetic dialogue dataset as a new benchmark; 2) we show that training over this dataset can improve the performance of an end-to-end dialogue system on empathetic dialogue.

\section{Related Work}

\paragraph{Emotion data} Crafting our dataset requires deciding what set of emotions the models should be capable of reacting to. 
Multiple schemas have attempted to organize the spectrum of emotions, from a handful of basic emotions derived from biological responses \citep{ekman1992argument,plutchik1984emotions} 
to larger sets of subtle emotions inferred from contextual situations \citep{SkerrySaxe}.
We incorporate emotions from multiple annotation schemas, noting that emotions merely
inferred from a situation are important in dialogue scenarios.  
There is a wide breadth of research in distributional representation approaches for many emotion classification tasks \citep{duppada2018seernet,park2018plusemo2vec,xu2018emo2vec,Mohammad2018SemEval2018T1} that build on
deep networks pretrained on large-scale weakly-labelled data such as emojis \citep{Deepmoji} or hashtags \citep{MohammadDataset}, gathered from public social media content published on Twitter.
The \textsc{semeval2019} EmoContext challenge also uses conversation data for detection of three basic emotions (`happy', `sad', and `angry')
over two turns of context
from Twitter exchanges
\citep{gupta2017sentiment}.  
We focus on personal conversations rather than using social media data 
to be closer to a context of a one-on-one conversation. 
Public social media content occurs in front of large ``peripheral audiences'' \citep{goffman1981forms} 
where
uncertainty as to how wide that audience is and the need for curated self-presentation \citep{goffman1959presentation} have been
shown to lead to different choices of subject matters
compared to private messaging, with people
sharing more intense and negative emotions
through private channels \citep{bazarova2015social,litt2014awkward}. 
In this work, we generate a more balanced
coverage of emotions than would appear in public social
media content, using a domain that is closer
to our ultimate goal of training a model for conversation that can respond to any emotion.

\paragraph{Controllable language generation} Several other works have focused on controlling the emotional content of a text response either through a manually specified target \citep{Mojitalk,Zhou17,Sentigan,Hu2017TowardCG,huang2018automatic} or through a general term to encourage higher levels of affect \citep{asghar2018affective}, with evaluations focused on matching a predetermined desired emotion rather than empathetic responding. \citet{Politeness} generate responses conditioned on
a specified politeness setting (polite, rude or neutral). 
\citet{huber2018emotional} investigate how to respond to emotions detected from an image.
Our work 
focuses on empathetic responses that are appropriate to signals inferred purely from text rather than conveying a pre-specified emotion.

\paragraph{Related chit-chat data} Several works have attempted to make chit-chat dialogue models more engaging by grounding them in personal contexts \citep{li2016persona,personachat,mazare2018training}, 
focusing on personal facts (``I am from New York''). 
Another interesting resource is the \textsc{DailyDialog} (DD) dataset \citep{DailyDialog}, which comprises about 13k dialogues obtained by crawling educational websites intended for learners of English and 
also has emotion label annotations.  Many of the dialogues are focused on topics for ESL learners (ordering from a restaurant, asking for directions, introductions, etc), but only $\approx 5 \%$ of the utterances have a label other than ``none'' or ``happy''. 
Our task focuses explicitly on conversations about emotionally grounded personal situations, and considers a richer, evenly distributed set of emotions.  We also introduce an explicit single \textit{listener} in the conversation who is reacting to the situation being described in an empathetic way, to make the setting as close as possible to our desired goal of a one-on-one empathetic conversation.

\begin{figure}
    \centering
    \includegraphics[trim={4.3cm 0cm 12.25cm 0cm},clip,width=.48\textwidth]{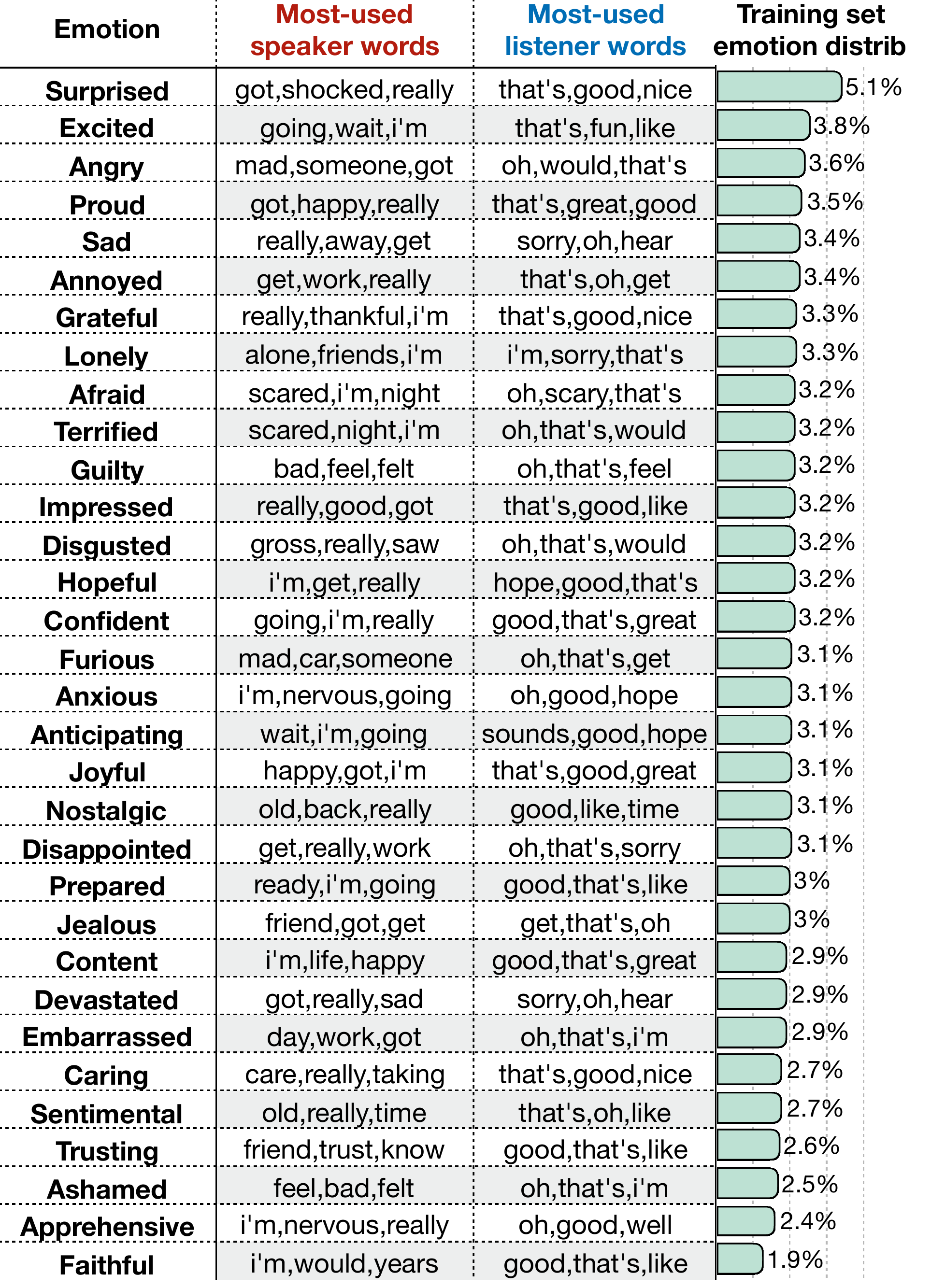}
    \caption{Distribution of conversation labels within \datasetname{} training set and top 3 content words used by speaker/listener per category. }
    \label{promptdistrib}
\end{figure}

\section{Talking about Personal Situations}

We consider an open-domain one-on-one conversational setting where two people are discussing a situation that happened to one of them, related to a given feeling. We collect around 25k conversations using the following format.

\paragraph{Emotional situation grounding} Each conversation is grounded in a situation, which one participant writes about in association with a given emotion label.
We consider 32 emotion labels, listed in Figure~\ref{promptdistrib}, which we chose by aggregating labels from several emotion prediction datasets \citep{ISEAR,SemEval2007,SkerrySaxe,DailyDialog, MohammadDataset}.  These emotion labels cover a broad range of positive and negative emotions. 
Our goal in providing a single emotion label is to have a situation strongly related to (at least) one particular emotional experience, though we note that some emotions may be very closely related\footnote{Researchers could merge similar emotions, like "afraid" and "terrified", to get coarser labels, if desired.} and additional related emotions may be invoked in a given conversation.

\paragraph{Speaker and listener} The person who wrote the situation description (\textit{Speaker}) initiates a conversation to talk about it. The other conversation participant (\textit{Listener}) becomes aware of the underlying situation through what the Speaker says and responds. Speaker and Listener then exchange up to 6 more turns.
We include two example conversations from the training data in Figure~\ref{dataex} and ten more in Table~\ref{moreex} in the Appendix.
The models discussed below are tested in the role of \textit{Listener} responding to the Speaker. Neither the situation description written by the Speaker nor the emotion label is given to the models (just as they were not given to the Listener during dialogue collection). 
Our data could also be used to generate conversations for the Speaker conditioned on the situation description though we leave this for future work.

\paragraph{Collection details}
We collected crowdsourced dialogues using the ParlAI platform \citep{miller2017parlai} to interact with Amazon Mechanical Turk (MTurk), hiring 810 US workers. 
A pair of workers are asked to (i) select an emotion word each and describe a situation when they felt that way, and to (ii) have a conversation about each of the situations, as outlined below.  Each worker had to contribute at least one situation description and
one pair of conversations: one as Speaker about the situation they contributed, and one as Listener about the situation contributed by another worker.  They were allowed to participate in as many  hits as they wanted for the first $\sim$10k conversations, then we limited the more ``frequently active'' workers to a maximum of 100 conversations. The median number of conversations per worker was 8, while the average was 61 (some workers were more active contributors than others). To ensure quality, we manually checked random subsets of conversations by our most-frequent workers.

\paragraph{Task set-up}
In the first stage of the task, workers are asked to describe in a few sentences a situation based on a feeling label. 
We ask the workers to try to keep these descriptions between 1-3 sentences. The average response is 19.8 words.
In the second stage, two workers are paired and asked to have two short chats with each other.  In each chat, one worker (\emph{speaker}) starts a conversation about the situation they previously described, and the other worker (\emph{listener}) responds.  Neither can see what the other worker was given as emotion label or the situation description they submitted, so they must respond to each others' stories based solely on cues within the conversation. Each conversation is allowed to be 4-8 utterances long (the average is 4.31 utterances per conversation).  The average utterance length was 15.2 words long.

\paragraph{Ensuring balanced emotion coverage}
After the first few initial rounds of data collection, we forced workers to select an emotion that among three emotion labels that 
 had been the least chosen overall so far if it was their first
 time working on the task. If they had already performed
 the task, the offered emotion labels were among those that
 they had chosen the least often before.
Given that a conversation model trained for empathetic responding needs to be able to handle emotions even if they are less frequent, we opted for this balancing procedure to make training for these categories easier, while still allowing for some measure of choice for workers.  As shown in Figure ~\ref{promptdistrib}, the distribution of emotion label prompts is close to evenly distributed, with a few that are selected slightly more/less often.

\paragraph{\datasetname{} dataset statistics}

The resulting dataset comprises 
24,850 conversations about a situation description, gathered from 810 different participants, which are publicly available through the ParlAI framework\footnote{https://parl.ai/} and for direct download with accompanying code\footnote{https://github.com/facebookresearch/EmpatheticDialogues}.
We split the conversations into approximately 80\% train, 10\% validation, and 10\% test partitions.  To prevent overlap of discussed situations between partitions, 
we split the data so that all sets of conversations with the same speaker providing the initial situation description would be in the same partition.  The final train/val/test split was 19533 / 2770 / 2547 conversations, respectively.   We include ten examples from our training set in Appendix Section~\ref{sec:examples}.

\section{Empathetic Response Generation}

\begin{figure}
    \centering
    \includegraphics[trim={5.45cm 16.75cm 17.cm 2.25cm},clip,width=.485\textwidth]{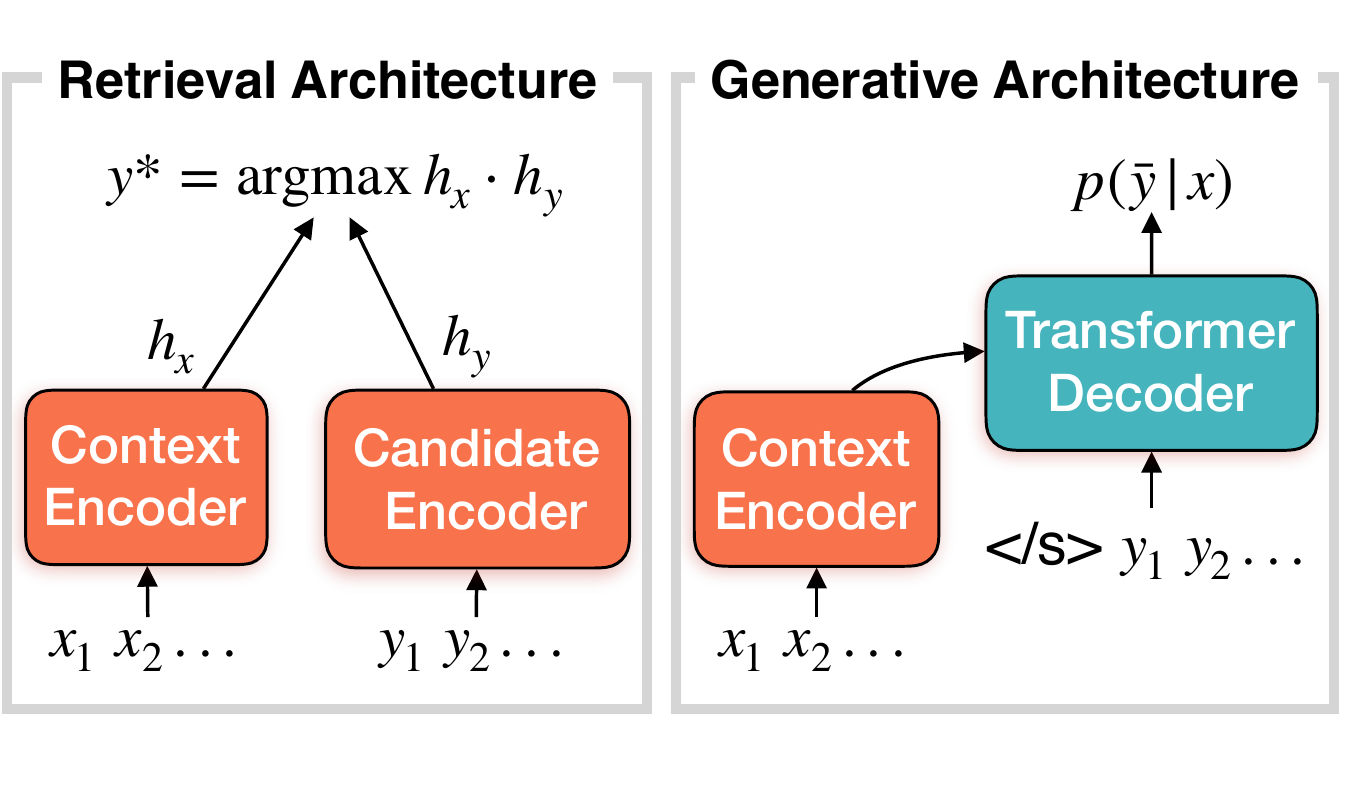}
    \vspace*{-7mm}
    \caption{Dialogue generation architectures used in our experiments. The context of concatenated previous utterances is tokenized into $x_1, x_2, \cdots$, and encoded into vector $h_x$ by the context encoder. \textit{Left:} In the retrieval set-up, each candidate $y$ is tokenized into $y_1, y_2, \cdots$ and encoded into vector $h_y$ by the candidate encoder.
    The system outputs the candidate $y^*$ that maximizes dot product $h_x \cdot h_y$. \textit{Right:} In the generative set-up,
    the encoded context $h_x$ is used as input to the decoder to generate start symbol $\texttt{</s>}$ and tokens $y_1, y_2, \cdots$. The model is trained to minimize the negative log-likelihood of target sequence $\bar{y}$ conditioned on context. }
    \vspace*{-2mm}
    \label{fig:arch1}
\end{figure}

This section shows how ED can 
be used as a benchmark to gauge the ability of a model to respond in an empathetic way, and as a training resource to make generic chitchat models more empathetic. We also examine different ways existing models can be combined to produce more empathetic responses.
We use ED dialogues to train and evaluate models in the task of generating conversation responses in the \textit{Listener} role. To emulate a normal conversation, the model has access to previous utterances in the dialogue, but not to the emotion word prompt (e.g., ``proud''), nor to the situation description generated by the Speaker.
Given a dialogue context $x$ of $n$ previous conversation utterances concatenated and tokenized as $x_1, \cdots, x_m$, followed by a target response $\bar{y}$, our models are trained to maximize the likelihood $p(\bar{y}|x)$ of producing the target response. We investigate both generative and retrieval-based settings \citep{lowe2016evaluation} as described in Figure~\ref{fig:arch1}.

\subsection{Base Architecture}
We base our models on Transformer networks \citep{TransformerNets}, which have proven successful in machine translation and dialogue generation tasks \citep{personachat,mazare2018training}. 

\paragraph{Retrieval-based}In the retrieval-based set-up, the model is given a large set $Y$ of candidate responses and picks the ``best'' one, $y^*$.
We first experiment with the retrieval Transformer-based architecture from \citet{Yangetal2018}: two Transformer encoders separately embedding the context, $x$, and candidates, $y \in Y$ ,  as $h_x$ and $h_y$, respectively. We also experiment with BERT \cite{bertppr} as base architecture 
to encode candidates and contexts, using the final hidden vector from BERT as the $h_x$ or $h_y$ encodings. The model chooses a candidate utterance according to a softmax on the dot product: $ h_x \cdot h_y$. 
We minimize the negative log-likelihood of selecting the correct candidate. 
At training time, we use all of the utterances from the batch as candidates, with a large batch size of 512 to give the model more negative examples (except for BERT for which a batch size of 256 was used).
At inference time, we experiment with three sets of candidate utterances for the model to choose from: all of the response utterances in the ED training set ($Y^{ED}$), all the utterances in the DailyDialog \citep{DailyDialog} training set ($Y^{DD}$), and a million utterances from a dump of 1.7 billion Reddit (R) conversations ($Y^{R}$).  

\paragraph{Generative} In the generative set-up, we use the full Transformer architecture \citep{TransformerNets}, consisting of
an encoder and a decoder.
The Transformer decoder uses the encoder output to predict a sequence of words $y$, and is trained to minimize the negative log-likelihood of
the target sequence $\bar{y}$. At inference time, we use diverse beam search from \citet{DiverseBS}.

\paragraph{Training details} Models are pretrained on predicting replies from a dump of 1.7 billion Reddit conversations, starting either from scratch for the Transformer architectures, or from the BERT$_{base}$ model released by \citet{bertppr} for the BERT-based architectures.\footnote{We used the Hugging Face PyTorch implementation of BERT at https://github.com/huggingface/pytorch-transformers. We experimented with directly fine-tuning BERT on ED without first training on Reddit conversations, but this did not perform as well.} Pretrained models without any fine-tuning on ED will be referred to as ``Pretrained'' hereafter.  
We
limit the maximum number of word tokens in the context and response to be 100 each.
The Transformer networks used in most experiments have the same base architecture (four layers and six transformer heads) and are trained the same way as in \citet{mazare2018training}. We also experiment with a larger architecture of five layers (denoted as "Large"), and BERT retrieval models, that are allowed to train for much longer (see training times in Table~\ref{table:capacity}).\footnote{While the models had not fully converged when we stopped training, we trained the Pretrained models for a few iterations more than the corresponding Fine-Tuned models, to ensure that any observed improvement was due to the data used for fine-tuning and not
the extra training time.}
For all models, we keep the version that has the lowest loss on the validation set.  For the Transformer models, we use 300-d word embeddings pretrained on common-crawl data using fastText~\citep{grave2018learning}, and for the BERT models, we use 768-d word embeddings pretrained on BooksCorpus and English Wikipedia \citep{bertppr}. More training details are provided in Appendix~\ref{sec:training}.

\subsection{Leveraging the Training Data from ED}
\label{sec:unsup}
A retrieval-based model relies on candidates. ED data  was explicitly collected with instructions to be empathetic,
in a one-on-one setting, which is not the case of the Reddit
conversation data used for pretraining, and these domain candidates may be better suited to empathetic responding than generic conversation utterances. Thus, we experiment with incorporating ED training candidates into the pool used at inference time by pretrained retrieval-based models, with no fine-tuning on ED.
For retrieval-based and generative models, 
  we also experiment with fine-tuning pretrained models to predict the next utterance over ED with a context window of four previous utterances, which is the average length of a conversation in our dataset. These models are referred to as ``\basename{}'' models. This fine-tuning is conducted until convergence for all architectures except those referred to as ``Pretrained''.
\begin{figure}
    \centering
    \includegraphics[trim={9.75cm 18.3cm 18.5cm 2.5cm},clip,width=.382\textwidth]{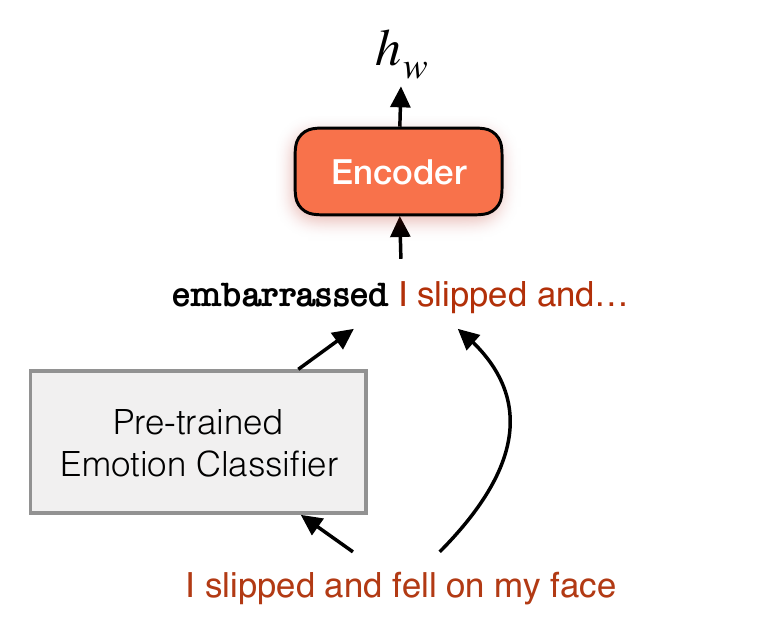}
    \vspace*{-2mm}
    \caption{
    Incorporating additional supervised information, here from an emotion classification task. 
    An input sequence (either a dialogue context or a candidate) is run through a pre-trained classifier, and the top $k$ output labels are prepended to the sequence, which is then run through the corresponding (context or candidate) encoder to output a hidden representation $h_w$ (either $h_x$ or $h_y$) as in the base setting.
    }
    \vspace*{-2mm}
    \label{fig:arch2}
\end{figure}

\begin{table*}[]
\begin{center}
\begin{tabular}{@{\extracolsep{5pt}}@{}lrrr@{\hspace*{2mm}}rr@{\hspace*{2mm}}rrr@{}}
 & \multicolumn{3}{c}{Retrieval} & \multicolumn{2}{c}{Retrieval w/ BERT} &\multicolumn{2}{c}{Generative} \\ 
\cline{2-4}\cline{5-6}\cline{7-8}\\
Model &\begin{tabular}[c]{@{}c@{}}Candidate \\ Source\end{tabular}
&P@1,100 &\begin{tabular}[c]{@{}c@{}}AVG\\ BLEU\end{tabular}& P@1,100 &\begin{tabular}[c]{@{}c@{}}AVG\\ BLEU\end{tabular}& PPL &\begin{tabular}[c]{@{}c@{}}AVG \\BLEU\end{tabular} \\ \midrule
Pretrained & R &-& 4.10 &-&4.26& 27.96 & 5.01 \\
 & ED & 43.25 & 5.51 &49.94&5.97& - & - \\
\basename{}& ED &  \textbf{56.90} &5.88 &65.92&\textbf{6.21}&\textbf{21.24} & \textbf{6.27} \\
 &ED+DD &-& 5.61 &-&-& - & - \\
 &ED+DD+R &-& 4.74 &-&-& - & - \\
{EmoPrepend-1} & ED & 56.31 &5.93&\textbf{66.04}&6.20& 24.30 & 4.36 \\
{TopicPrepend-1}  & ED & 56.38 &\textbf{6.00}&65.96&6.18& 25.40 & 4.17 \\
\bottomrule
\end{tabular}
\end{center}
\caption{Automatic evaluation metrics on the test set.
Pretrained: model pretrained on a dump of 1.7 billion \textsc{Reddit} conversations (4-layer Transformer architecture, except when specified BERT).
\basename{}: model fine-tuned over the \datasetname{} training data (Sec.~\ref{sec:unsup}).
EmoPrepend-1, Topic-Prepend1: model incorporating supervised information from an external classifiers, as described in Sec.~\ref{sec:explicit}.
Candidates come from \textsc{Reddit} (R), \datasetname{} (ED), or \textsc{DailyDialog} (DD). 
P@1,100: precision retrieving the correct test candidate out
of 100 test candidates.
AVG BLEU: average of BLEU-1,-2,-3,-4. PPL: perplexity.
All automatic metrics clearly improve with in-domain training on utterances (\basename{} vs. Pretrained), other metrics
are inconsistent. 
\textit{Bold: best performance for that architecture.}
}
\label{table:automatic}
\end{table*}

\subsection{Adding Information from External Predictors}
\label{sec:explicit}

Many existing models have been pretrained on supervised tasks that
may be relevant to empathetic responding. 
Combining these models with the representations from our
base architecture may reap benefits from previous training time
and external training data without having to redo the work or requiring access to that data, which may matter to practitioners.
Note that this may considerably augment the effective 
capacity of the resulting models, as well as the total
amount of training data used overall, but 
our goal here is to get an empirical sense of how robust performance improvement is to variations in architecture set-up or supervision domain.
We experiment with adding supervised information from two prediction tasks: emotion detection, which is more closely relevant to our task, and topic detection, which may also be useful in crafting relevant replies.\footnote{We considered multitask or feature concatenation set-ups, but they did not provide consistent improvements. These experiments are included in Appendix~\ref{sec:addexp}.}

\paragraph{Prepending Top-k Predicted Labels}
This set-up (Fig. \ref{fig:arch2}
), \textsc{Prepend-1}, is a very simple way to add
supervised information to data, requires no architecture modification, and can be used with black-box classifiers.
The top predicted label\footnote{We only discuss prepending the top predicted label here, but also experimented with top-3 and top-5 models, with similar result patterns, shown in Appendix~\ref{sec:addexpres}.} from the supervised classifier
is merely prepended to the beginning of the token sequence as encoder input, as below:

{
\hspace*{.2in}\textbf{Original}:``I finally got promoted!'' \\
\hspace*{.2in} 
\textbf{Prepend-1}:``\texttt{proud} I finally got promoted!''
}

Similar methods have been used for 
controlling the style of generated text (e.g. \citealp{Politeness}). 
Here, we use a fastText model \citep{fasttext}
as prediction architecture.
Both the context and the candidates are run through the classifier
and receive prepended labels. Fine-tuning is conducted similarly as
before, but using these modified inputs.
We use two external sources of information.
To provide emotion signal, we train a classifier to predict the emotion label from the description of the situation written by the Speaker before the dialogue for the training set dialogues of ED (\textsc{EmoPrepend-1}).\footnote{We also experimented with training the classifier on the utterances themselves, with similar results.}
To gauge whether supervision from a more distant task
would still be helpful, we also experiment with a classifier trained on the 20-Newsgroup dataset \citep{joachims1996probabilistic}, for topic classification (\textsc{TopicPrepend-1}).

\begin{table*}[tb]
\begin{tabular}{llllll}
\toprule
                           &    Model          & Candidate   & Empathy &  Relevance& Fluency \\
\midrule
\multirow{7}{*}{Retrieval} & {\it Pre-trained} & R  & $2.82\pm0.12$& $3.03\pm0.13$&$4.14\pm0.10$ \\
                           &              & R+ED & $3.16\pm0.14$&	$3.35\pm0.13$&	$4.16\pm0.11$\\
                           &              & ED & $3.45\pm0.12$&	$3.55\pm0.13$&	$4.47\pm0.08$\\

                           &  Fine-tuned   & ED & ${\bf3.76\pm0.11}$&$3.76\pm0.12$&$4.37\pm0.09$ \\
                           & EmoPrepend-1  & ED&$3.44\pm0.11$&$3.70\pm0.11$&$4.40\pm0.08$\\
                           & TopicPrepend-1&ED&$3.72\pm0.12$&${\bf 3.91\pm0.11}$&${\bf 4.57\pm0.07}$

\\

\hline
\multirow{7}{*}{Retrieval w/ BERT} & {\it Pre-trained }& R  & $3.06\pm0.13$&$3.29\pm0.13$&$4.20\pm0.10$
  \\
                           &              & R+ED & $3.49\pm0.12$&	$3.62\pm0.12$&	$4.41\pm0.09$\\
                           &              & ED & $3.43\pm0.13$&	$3.49\pm0.14$&	$4.37\pm0.10$
\\

                           &   Fine-tuned           & ED & $3.71\pm0.12$&$3.76\pm	0.12$&	$4.58\pm0.06$
 \\                           & EmoPrepend-1  & ED &$3.93\pm0.12$&$3.96\pm0.13$&$4.54\pm0.09$\\
                           & TopicPrepend-1  & ED & ${\bf 4.03\pm0.10}$&${\bf 3.98\pm0.11}$&${\bf 4.65\pm0.07}$\\

\hline
\multirow{3}{*}{Generative}& {\it Pre-trained} & -- &$2.31\pm0.12$&	$2.21\pm0.11$&	$3.89\pm0.12$
 \\
                           & Fine-Tuned  &  --  & ${\bf 3.25\pm0.12}$&	${\bf 3.33\pm0.12}$&	$4.30\pm0.09$ \\
                           &EmoPrepend-1  &  -- &$3.16\pm0.12$&$3.19\pm0.13$&$4.36\pm0.09$\\

                           &TopicPrepend-1 & -- &$3.09\pm0.13$&$3.12\pm0.13$&${\bf 4.41\pm0.08}$
\\

\hline
{\it Gold Response} &--&--&$4.19\pm0.10$&$4.55\pm0.07$&$4.68\pm0.06$\\
\bottomrule
\end{tabular}
\caption{Human ratings. Fine-tuning on \datasetabbrev{} and using ED candidates generally improves scores, especially on Empathy, with minimal retraining. Additional external supervision (Prepend) improves the Empathy and Relevance scores for BERT-based models. Bold: best score for that group. Italics: reference model for the group.}
\label{table:humaneval}
\end{table*}

\section{Experimental Evaluation}

We evaluate the models on their ability to reproduce the Listener's portion of the conversation (i.e. the ability to react to someone else's story).  We use both automated metrics and human evaluation to score each model's retrievals/generations. Human evaluation is important, as automated metrics don't always correlate with human judgments of dialogue quality \citep{HowNotToEval}, but we provide automated metrics to give a sense of how well they
align with human judgment on this task.

\paragraph{Automated metrics (Table~\ref{table:automatic})} For both retrieval and generative systems, we compute BLEU scores \citep{papineni2002bleu} for the model response, comparing against the gold label (the actual response), following
the practice of earlier work in dialogue generation \citep{wen2015semantically,li2015diversity,li2016persona}.
For the generative systems, we additionally report perplexity of the actual gold response. 
For the retrieval-based systems, we further compute $\texttt{p@1,100}$, the accuracy of the model at choosing the correct response out of a hundred randomly selected examples in the test set.  When we compute $\texttt{p@1,100}$, the actual response is included in the candidates, unlike inference from the retrieval systems for all other metrics, which only uses training utterances as candidates.

\paragraph{Human ratings (Table~\ref{table:humaneval})} We ran crowdsourcing tasks on MTurk (further details in Appendix~\ref{sec:evaltask}). Participants were given a model's output for a randomly selected test set example and asked to score different aspects of the model. The rating task provides a means of comparing aspects of responses, and we ask raters specifically about whether the response is acknowledging the conversation partner's feelings. We collected at least 100 ratings per model and asked about three aspects of performance, all rated on a Likert scale (1: not at all, 3: somewhat, 5: very much):
\\
\hspace*{.2cm}\textbf{ Empathy/Sympathy:} did the responses show understanding of the feelings of the person talking about their experience? \\
\hspace*{.2cm}\textbf{ Relevance:} did the responses seem appropriate to the conversation?  Were they on-topic? \\
\hspace*{.2cm}\textbf{ Fluency:} could you understand the responses?  Did the language seem accurate?
\\

\begin{table*}[tb]
\begin{center}
\begin{tabular}{llllll}
\toprule
  & Model & Params, resources, train examples & Emp & Rel & Fluent \\ \midrule
\multirow{6}{*}{Retrieval} &  Pretrained-R &84.3M, 2.5 days, 8GPUs, 1.7B&  2.8 & 3.0 & 4.1 \\
 &Pretrained-ED & same , same, same&   3.5 & 3.6 & 4.5 \\
 &\basename{} & same , + 0.5 hour, 1 GPU, +22.3k&   3.8 & 3.8 & 4.4 \\
 \cline{2-6}
 &Pretrained-Bert-R&217M, 13.5 days, 8GPUs , 1.7B &3.1&3.3&4.2\\
 &Pretrained-Bert-ED&same, same, same &3.4&3.5&4.4\\
 &\basename{}-Bert&same, +1hour, 8GPUs, +22.3k&3.7&3.8&4.6\\
 \midrule
\multirow{4}{*}{Generative} &Pretrained &85.1M, 2 days, 32 GPUs, 1.7B&   2.3 & 2.2 & 3.9 \\
  &\basename  & same , +1 hour, 1 GPU, +22.3k&  3.3  & 3.3  & 4.3 \\
  \cline{2-6}
  &Pretrained-Large &86.2M, 2.5 days, 32 GPUs, 1.7B&  2.8 & 3.0 & 4.0 \\
  &\basename{}-Large & same , +0.5 hour, 1 GPU, +22.3k& 3.6 & 3.6 & 4.5 \\
 \hline
\end{tabular}
\end{center}
\caption{Training
resources for different models, with human 
ratings
for empathy (Emp), relevance (Rel) and fluency (Fluent). Retrieval-based models use reply candidates from the ED training set (ED) or from Reddit (R). Resource comparisons are relative to the first row of each group. Fine-tuning on \datasetabbrev{} improves all scores (except for Fluency in one case) while requiring minimal additional training resources.
SEM is approximately 0.1}
\label{table:capacity}
\end{table*}

\begin{table*}[tb]
\begin{center}
\begin{tabular}{p{2.5cm}lllp{6cm}}
\toprule

Context & Setup&Model&Cand & Generated Response \\ \midrule
\multirow{2}{2.4cm}{Someone came to my door with a gun the other day.} &
\cellcolor{LightGray}Gen&\cellcolor{LightGray}Pretrained&\cellcolor{LightGray}- &\cellcolor{LightGray}I'm not sure if you're joking or not.\\
 &\cellcolor{Gray}Gen&\cellcolor{Gray}\basename{}&\cellcolor{Gray}- &\cellcolor{Gray}Oh no! What did you do?\\
 &\cellcolor{LightGray}Gen&\cellcolor{LightGray}EmoPrepend&\cellcolor{LightGray}-&\cellcolor{LightGray}Oh no! Did you have a weapon?\\
 &\cellcolor{Gray}Gen&\cellcolor{Gray}TopicPrepend&\cellcolor{Gray}-&\cellcolor{Gray}I'm glad to hear that.\\
\midrule

\multirow{9}{2.4cm}{My son failed his exams! Can you believe it! I don't know what to do!}&\cellcolor{LightGray}Ret-BERT&\cellcolor{LightGray}Pretrained&\cellcolor{LightGray}R &\cellcolor{LightGray}Hopefully his mistakes can help you know what not to do on the tests.\\ 
&\cellcolor{Gray}Ret-BERT&\cellcolor{Gray}Pretrained&\cellcolor{Gray}ED &\cellcolor{Gray}Oh no! I ' m so sorry! I hope you could pull your grade up!\\

&\cellcolor{LightGray}Ret-BERT&\cellcolor{LightGray}\basename{}&\cellcolor{LightGray}ED &\cellcolor{LightGray}Oh no! Did he not study? \\
&\cellcolor{Gray}Ret-BERT&\cellcolor{Gray}EmoPrepend&\cellcolor{Gray}ED &\cellcolor{Gray}Im sorry about that! Did your son start school already?\\
&\cellcolor{LightGray}Ret-BERT&\cellcolor{LightGray}TopicPrepend&\cellcolor{LightGray}ED &\cellcolor{LightGray}Oh no! Does he have the option of taking it again? \\
\bottomrule
\end{tabular}
\end{center}
\caption{Examples of model responses. Shading is provided for better legibility.
}
\label{table:modelout}
\end{table*}
\subsection{Results}
\paragraph{Pretrained models baseline}
Pretrained conversation models are rated poorly by humans for empathy when the candidates are retrieved from 
Reddit utterances or when a generative model is used  (Table~\ref{table:humaneval}). Higher ratings with models based on BERT or larger Transformer models show that increasing the capacity makes the models seem more empathetic, but still remain far
from human performance, while being considerably more onerous to train (Table~\ref{table:capacity}).\footnote{Results on larger retrieval-based Transformer models in Table~\ref{table:capacity_full} of the Appendix show the same pattern.}

\paragraph{Using \datasetname{} for candidate selection}
Table~\ref{table:automatic} shows that merely using the pool of candidates from the training set of \datasetabbrev{} improves the BLEU scores of retrieval models.

Using candidates from our dataset also substantially improves the performance of pre-trained retrieval models on all human metrics, particularly the Empathy subscore of most interest to us (Table~\ref{table:humaneval}).

\paragraph{Using \datasetname{} for fine-tuning}
Additionally, fine-tuning to predict conversation responses on our data improves all automated metrics (Table~\ref{table:automatic}). 
While fine-tuning on ED data improves performance on predicting the next ED utterance, this may come at the expense of performance when predicting next utterance in other corpora. To measure this, we compared automated metrics on next utterance prediction with pre-trained models and models fine-tuned using ED data (for our base and larger retrieval-based Transformer models) when predicting on 
\textsc{DailyDialog} and \textsc{Reddit} (drawing both context and candidates from the same corpus).
Compared to the 12-14\% P@1,100 increase measured with ED (see Tables~\ref{table:automatic} and \ref{tab:automatic_full}), fine-tuning on ED leads to a 5-7\% increase on DD, and a 2-3\% decrease on R.\footnote{Numbers for these datasets are included in Table~\ref{tab:ddr} of the appendix.} For all three datasets, fine-tuning increases AVG BLEU by 0.2 to 0.5.
The slight decrease of performance on R is not surprising because the pre-trained model was trained directly on Reddit predictions. But, the improvement on DD is an encouraging sign that improvements from fine-tuning on ED may generalize to other conversation datasets.

Fine-tuning on the \datasetabbrev{} data also generally improves human metrics on the ED task, in both retrieval and generative set-ups (Table~\ref{table:humaneval}).

\paragraph{Augmenting conversation models with external pretrained classifiers}
Automated and human evaluations suggest that prepending emotion or topic predictions may boost perfomance of high-capacity models based on BERT (but not the smaller models), with Empathy ratings close to approaching human performance.

More extensive experiments with large models would be required to confirm that larger capacity makes additional external supervision effective for this task.

\paragraph{Resources and capacity}
Table~\ref{table:capacity} quantifies resource and parameter usage for several models and set-ups, including

a larger Transformer generative model (5 layers instead of 4) and
 BERT-based architectures with substantially more parameters that require longer training.
Using ED candidates in pretrained retrieval models, or fine-tuning pretrained conversation models on
ED data makes smaller models perform better than larger ones with minimal increase in resource usage.

\section{Conclusion}
We introduce a new dataset of 25k dialogues grounded in situations prompted by specific emotion labels.  
Our experiments show that using this dataset to provide retrieval candidates or fine-tune
conversation models leads to responses that are evaluated as more empathetic. 
how to integrate empathetic responding into more general dialogue when, for example, the needs for empathy have to be balanced with staying on topic or providing information. 
We hope that our results and dataset will stimulate more research in the important direction of making dialog systems more empathetic.

\section*{Acknowledgments}
We thank the anonymous reviewers for insightful feedback and suggestions. This material is based, in part, upon work supported by the National Science Foundation Graduate Research Fellowship Program under Grant No. DGE-1256082.

\bibliography{aclbib}
\bibliographystyle{acl_natbib}

\clearpage
\appendix

\begin{table*}
\begin{center}

{\footnotesize
\resizebox{\textwidth}{!}{
\begin{tabular}{p{2.65in}|p{3.25in}}
\hspace*{-3mm}
\begin{tabular}[t]{p{2.6in}} \textbf{Label: Content}\\ \textbf{Situation:} Speaker felt this when...\\``eating my favorite meal makes me happy.'' \\ \textbf{Conversation:} \\\vspace*{-2.25mm} {\color{red} Speaker}: i am at my best when i have my favorite meal.\\\vspace*{-2.25mm} {\color{blue} Listener}: nice\\\vspace*{-2.25mm} {\color{red} Speaker}: i love enchiladas\\\vspace*{-2.25mm} {\color{blue} Listener}: really?\\\vspace*{-2.25mm} {\color{red} Speaker}: yes. enchiladas for the win!\\ \\ \textbf{Label: Proud}\\ \textbf{Situation:} Speaker felt this when...\\``I was proud when my brother finished college. He worked so hard at it'' \\ \textbf{Conversation:} \\\vspace*{-2.25mm} {\color{red} Speaker}: I was proud of my brother when he finished school. He worked so hard at it\\\vspace*{-2.25mm} {\color{blue} Listener}: Nice, tell him congrats. What did he major in?\\\vspace*{-2.25mm} {\color{red} Speaker}: It was English\\\vspace*{-2.25mm} {\color{blue} Listener}: He should become an English teacher1\\ \\ \textbf{Label: Joyful}\\ \textbf{Situation:} Speaker felt this when...\\``I have had a great week!'' \\ \textbf{Conversation:} \\\vspace*{-2.25mm} {\color{red} Speaker}: I have had a great start to my week!\\\vspace*{-2.25mm} {\color{blue} Listener}: That's great. Do you think the rest of the week will be as great?\\\vspace*{-2.25mm} {\color{red} Speaker}: I hope so!  It looks promising!!  \\\vspace*{-2.25mm} {\color{blue} Listener}: Lucky you. Are you always a positive person or it's just been an amazing week really?\\\vspace*{-2.25mm} {\color{red} Speaker}: haha.  Kind of both.  And also probably too much coffee to start my shift tonight\\ \\ \textbf{Label: Terrified}\\ \textbf{Situation:} Speaker felt this when...\\``I got home for lunch and found a bat outside on my front porch.'' \\ \textbf{Conversation:} \\\vspace*{-2.25mm} {\color{red} Speaker}: I got home for lunch and found a bat outside on my front porch. It probably has rabies. Bats shouldn't be out during the day.\\\vspace*{-2.25mm} {\color{blue} Listener}: Doesn't rabies cause sensativity to light? Either way I would freak out...\\\vspace*{-2.25mm} {\color{red} Speaker}: It can but, it also causes anmails to behave erratically... like bats wadering around in the middle of the day.\\\vspace*{-2.25mm} {\color{blue} Listener}: Oh yeah, gotcha. I really don't like animals that are small and move quickly\\\vspace*{-2.25mm} {\color{red} Speaker}: Generally yes.\\ \\ \textbf{Label: Anticipating}\\ \textbf{Situation:} Speaker felt this when...\\``I cant wait to go on my end of summer trip'' \\ \textbf{Conversation:} \\\vspace*{-2.25mm} {\color{red} Speaker}: I cant wait to go on my end of summer trip in texas.\\\vspace*{-2.25mm} {\color{blue} Listener}: Sounds like fun. What you got planned ?\\\vspace*{-2.25mm} {\color{red} Speaker}: not really sure but im excited to just be invited\\\vspace*{-2.25mm} {\color{blue} Listener}: Got any family out there? Cousins perhaps\end{tabular}&
\begin{tabular}[t]{p{3.25in}} \textbf{Label: Terrified}\\ \textbf{Situation:} Speaker felt this when...\\``My brother jump scared me while I was out playing. It was crazy bad.'' \\ \textbf{Conversation:} \\\vspace*{-2.25mm} {\color{red} Speaker}: Just got scared to death.\\\vspace*{-2.25mm} {\color{blue} Listener}: Oh no. What happened?\\\vspace*{-2.25mm} {\color{red} Speaker}: My brother jumped scared me.\\\vspace*{-2.25mm} {\color{blue} Listener}: lol is he younger or older?\\ \\ \textbf{Label: Proud}\\ \textbf{Situation:} Speaker felt this when...\\``My little dog learned to sit!'' \\ \textbf{Conversation:} \\\vspace*{-2.25mm} {\color{red} Speaker}: I finally tough my new little puppy his first trick!\\\vspace*{-2.25mm} {\color{blue} Listener}: What trick did you teach him?\\\vspace*{-2.25mm} {\color{red} Speaker}: I tought him to sit for a treat, its so cute.\\\vspace*{-2.25mm} {\color{blue} Listener}: That is good, do you plan to teach him more tricks?\\ \\ \textbf{Label: Apprehensive}\\ \textbf{Situation:} Speaker felt this when...\\``I have to call my landlord about being late on the rent. I really don't want to have this conversation.'' \\ \textbf{Conversation:} \\\vspace*{-2.25mm} {\color{red} Speaker}: I have to make a dreadful phone call tomorrow\\\vspace*{-2.25mm} {\color{blue} Listener}: Oh no, about what?\\\vspace*{-2.25mm} {\color{red} Speaker}: I'm late on my rent and I need another week. I don't want to because my landlord isnt very nice\\\vspace*{-2.25mm} {\color{blue} Listener}: Oh no, I've been there done that too many times.\\\vspace*{-2.25mm} {\color{red} Speaker}: I don't want her to make a big deal\\ \\ \textbf{Label: Confident}\\ \textbf{Situation:} Speaker felt this when...\\``When my husband asked me about how to build a chicken coop I was able to give him a reply that was backed up by blueprints and research from the internet. '' \\ \textbf{Conversation:} \\\vspace*{-2.25mm} {\color{red} Speaker}: We recently got 9 chicks and we've been having to work on making them a coop! I had to do so much research but I think we finally have a place that they'll enjoy living when they aren't able to free range.\\\vspace*{-2.25mm} {\color{blue} Listener}: OHH! I Love chickens ! I have always wanted some. I have a duck! lol- What kind of chickens are they?\\\vspace*{-2.25mm} {\color{red} Speaker}: We currently have 2 Australorps, 3 Rhode Island Reds, 3 Barred Plymouth Rocks, and 1 Welsummer, but 4 of the 9 ended up being roosters. Ugh!\\\vspace*{-2.25mm} {\color{blue} Listener}: Oh man! They fight sometimes. I hope they aren't too bad about waking you up in the morning. Chickens can be very sweet though!\\\vspace*{-2.25mm} {\color{red} Speaker}: I love my little hens, especially one I've named Curly. The roosters might get replaced by hens though because the crowing is so frustrating! \\  \\ \textbf{Label: Surprised}\\ \textbf{Situation:} Speaker felt this when...\\``I got a lottery ticket while I was at work today. I won \$100 on the scratch off. I was shocked. I never win.'' \\ \textbf{Conversation:} \\\vspace*{-2.25mm} {\color{red} Speaker}: I won \$100 on a scratch off today. I was shocked. I never win.\\\vspace*{-2.25mm} {\color{blue} Listener}: Wow! How often do you play the lottery?\\\vspace*{-2.25mm} {\color{red} Speaker}: I usually go on our Tuesday break to buy one with coworkers.\\\vspace*{-2.25mm} {\color{blue} Listener}: Neat! Well that is a fantastic feat. Maybe you can win again sometime?\end{tabular}\\

\end{tabular}}}
\end{center}
\caption{10 random examples from \datasetname{} training set.}
\label{moreex}
\end{table*}

\section{Data Examples}
\label{sec:examples}
We include ten randomly selected dialogues from our training set in Table~\ref{moreex}.

\section{Human Evaluation Crowdsourcing Task}
\label{sec:evaltask}
Human evaluations were collected on MTurk. For the rating task, each worker was shown a set of 10 randomly subsampled examples from the test set, one after another, each from a different randomly selected model. The worker had to rate the empathy, relevance, and fluency of each example before moving onto the next one. At least 100 ratings were collected per model. 221 US workers participated in the rating task, and each had to perform a minimum of one set of 10 ratings.

\begin{table}[]
\begin{tabular}{@{}lllll@{}}
\toprule
                  & \multicolumn{2}{l}{P @1,100} & \multicolumn{2}{l}{BLEU} \\ \midrule
Model             & DD            & R            & DD          & R          \\ \midrule
Pretrained       & 39.04         & 58.95        & 6.65        & 1.43       \\
Fine-Tuned         & 44.58         & 56.25        & 7.14        & 1.64       \\\midrule
Pretrained-Large & 42.28         & 61.60        & 6.94        & 1.42       \\
Fine-Tuned-Large  & 48.96         & 58.71        & 7.42        & 1.73       \\ \bottomrule
\end{tabular}
\caption{Performance of the retrieval-based pretrained model and retrieval-based models fine-tuned on ED data for next utterance prediction in other datasets, with both context and candidates from the same dataset (R=Reddit, DD=DailyDialog).}
\label{tab:ddr}
\end{table}
\section{Next utterance prediction on other datasets}
We test how fine-tuning on ED data affects next utterance prediction on two external datasets (\textsc{Reddit} and \textsc{DailyDialog}).  In this experiment, we use both candidates and context from the DD or R data.  Results in Table~\ref{tab:ddr} show that performance on \textsc{DailyDialog} improves after fine-tuning on our data.

\section{Additional Experimental Details and Results}

\subsection{Training Details}
\label{sec:training}
We used Adamax for training throughout, and dropout was set to 0\% everywhere except for a 20\% dropout in the linear layer of the emotion-label term of the \textsc{Multitask} objective function (discussed below). A learning rate of $8\mathrm{e}{-4}$ was used for all four-layer Transformer models, following \citet{mazare2018training}. For the five-layer retrieval-based Transformer model (Pretrained-Large and Fine-Tuned-Large), the learning rate was selected by picking the best performing over the validation set, among values randomly sampled between between $5\mathrm{e}{-5}$ and $8\mathrm{e}{-4}$. When training the retrieval-based BERT model on Reddit and ED data, the learning rate was selected by picking the best performing over the validation set, among values randomly sampled between $6\mathrm{e}{-6}$ and $2\mathrm{e}{-4}$. For training BERT models on Reddit data, we also experimented with adding an additional Transformer layer after the output embedding of the BERT model, but this slightly degraded P@1,100 scores on the validation set. We used a learning rate of  $8\mathrm{e}{-5}$ for the five-layer generative Transformer models.

\subsection{Additional Experimental Set-Ups}
\label{sec:addexp}
We investigated a few additional approaches for incorporating supervised emotion or topic prediction in generating dialogue, but observed little performance improvement. Methods are described below.

\paragraph{Multitask with Emotion labels}
\label{sec:mt}
If the most appropriate response depends on 
some information for which supervision is available, e.g.,
the emotions at play, nudging the model to encode
this information could result in better performance.
We experimented with this by training the base architecture in the
 one-to-many style of multi-task learning that has been used for NLP seq2seq settings \citep{multitask}. In this set-up, 
 \textsc{Multitask}, we altered the objective function to also optimize for predicting the emotion label of the conversation to which the utterances being encoded belonged. We added to the context encoder a linear layer and softmax that predicted the emotion label from the context sentences.  The objective function was altered to be the average of the negative log-likelihood of predicting the next utterance $\bar{y}$ and the negative log-likelihood of the added linear layer being able to predict the correct emotion.
\paragraph{Prepend-3, Prepend-5}
We investigated whether Prepend models could be improved by adding the top-3/5 predicted emotion or topic labels by the classifier (rather than top-1).

\paragraph{Ensemble of Encoders}
We also investigated another approach for incorporating external predictors, which we report the results of in our extended results tables. In this set-up (\textsc{Ensem}), we augmented the encoders to incorporate 
latent representations from pretrained supervised architectures.
We replaced each of the encoders in our Transformer networks with an Ensemble encoder
, similar to a many-to-one style encoder-decoder architecture \citep{multitask}.  This encoder took the encoding $h_w$ from our basic Transformer encoder (either $h_x$ or $h_y$), already trained on our data, and concatenated it with the representation $h_c$ extracted from the inner layer of a classification
network. We used the penultimate layer of a deep emotion classifier.
The concatenated encodings were projected linearly to the dimension required by the decoder, whose architecture didn't change.  
When training the dialogue model,
we froze both
the base Transformer encoder 
and the pretrained classifier 
and trained
only the linear layers (and the decoder for generative systems).
We used emotion-related supervision from Emojis from Twitter, through the use of
the trained Deepmoji system \citep{Deepmoji} released by the authors, either as-is (\textsc{Ensem-DM}) or fine-tuned on the
situation descriptions of \datasetname{} (\textsc{Ensem-DM+}).

\subsection{Additional Experiments Results}
\label{sec:addexpres}
Automated and human evaluations for any additional experiments are in Tables \ref{table:automatic_supp_full} and \ref{table:human_supp_full}, respectively.  All of these model variations show improvements over the pre-trained models.  In some metrics, many of these models show slight improvements over the fine-tuned models, as well, though not as consistently, except for the larger BERT retrieval-based models.  While prepending top-1 or top-3 labels do not improve generative model scores, the results in Table~\ref{table:human_supp_full} suggest that multitask, prepend-5, and ensemble set-ups may improve the human evaluations of the fine-tuned generative model for empathy, but are too inconsistent to be conclusive without more corroborating experiments. 

\begin{table*}[]
\begin{center}
\begin{tabular}{@{\extracolsep{5pt}}@{}lrrr@{\hspace*{2mm}}rr@{\hspace*{2mm}}rrr@{}}
 & \multicolumn{3}{c}{Retrieval} & \multicolumn{2}{c}{Retrieval w/ BERT} &\multicolumn{2}{c}{Generative} \\ 
\cline{2-4}\cline{5-6}\cline{7-8}\\
Model &\begin{tabular}[c]{@{}c@{}}Candidate \\ Source\end{tabular}
&P@1,100 &\begin{tabular}[c]{@{}c@{}}AVG\\ BLEU\end{tabular}& P@1,100 &\begin{tabular}[c]{@{}c@{}}AVG\\ BLEU\end{tabular}& PPL &\begin{tabular}[c]{@{}c@{}}AVG \\BLEU\end{tabular} \\ \midrule

Pretrained       & R  & -     & 4.10 & -     & 4.26 & 27.96 & 5.01 \\
               & R+ED & -     & 4.96 & -     & 5.62 & -     & -    \\
                 & ED & 43.25 & 5.51 & 49.94 & 5.97 & -     & -    \\
\basename{}      & R  & -     & 3.85 & -     & 4.14 & -     & -    \\
               & R+ED & -     & 4.76 & -     & 5.39 & -     & -    \\
                 & ED & 56.90 & 5.88 & \textbf{65.92} & 6.21 & 21.24 & 6.27 \\
               &ED+DD & -     & 5.61 & -     & -    & -     & -    \\
             &ED+DD+R & -     & 4.74 & -     & -    & -     & -    \\
\midrule
{Pretrained-Large} & R  & -   & 4.16 & -     & -    & -     & -    \\
                 & ED & 47.58 & 5.78 & -   & -    & 23.64 & 6.31 \\
{\basename{}-Large} & ED & \textbf{60.44} & 6.01 & -  & -    & \textbf{16.55} & \textbf{8.06} \\
\midrule
{Multitask}      & ED & 55.73 & 6.18 & 65.90 & 6.17 & 24.07 & 5.42 \\
{EmoPrepend-1}   & ED & 56.31 & 5.93 & 66.04 & 6.20 & 24.30 & 4.36 \\
{EmoPrepend-3}   & ED & 55.75 & \textbf{6.23} & 65.85 & 6.14 & 23.96 & 2.69 \\
{EmoPrepend-5}   & ED & 56.35 & 6.18 & 64.69 & 6.21 & 25.40 & 5.56 \\
{TopicPrepend-1} & ED & 56.38 & 6.00 & 65.96 & 6.18 & 25.40 & 4.17 \\
{TopicPrepend-3} & ED & 55.44 & 5.97 & 65.85 & \textbf{6.25} & 25.02 & 3.13 \\
{TopicPrepend-5} & ED & 55.75 & 6.17 & 65.65 & 6.19 & 25.10 & 6.20 \\
{Ensem-DM}       & ED & 52.71 & 6.03 & -     & -    & 19.05 & 6.83 \\ 
{Ensem-DM+}      & ED & 52.35 & 6.04 & -     & -    & 19.10 & 6.77 \\

\bottomrule
\end{tabular}
\end{center}
\caption{
\label{tab:automatic_full}
Automatic evaluation metrics on the test set for full set of experimental setups.
Pretrained: basic Transformer model pretrained on a dump of 1.7 billion \textsc{Reddit} conversations.
\basename{}: model fine-tuned over the \datasetname{} training data.
Multitask: model trained with multitask loss function (predicting the emotion label).
EmoPrepend-1/3/5, TopicPrepend-1/3/5: model using top-k labels outputted by an external classifier as prepended tokens.
Ensem: model incorporating external classifiers by concatenating representations from deepmoji with the fine-tuned transformer representation.
Candidates come from \textsc{Reddit} (R) or \datasetname{} (ED).
P@1,100: precision retrieving the correct test candidate out
of 100 test candidates.
AVG BLEU: average of BLEU-1,-2,-3,-4. PPL: perplexity.
\textit{Bold: best performance in that column.}
}
\label{table:automatic_supp_full}
\end{table*}
\begin{table*}[tb]
\begin{tabular}{llllll}
\toprule
                           &    Model          & Candidate   & Empathy &  Relevance& Fluency \\
                           
\midrule
\multirow{18}{*}{Retrieval} 
& {\it Pretrained} & R  & $2.82\pm0.12$& $3.03\pm0.13$&$4.14\pm0.10$ \\
&              & R+ED & $3.16\pm0.14$&	$3.35\pm0.13$&	$4.16\pm0.11$\\
&              & ED & $3.45\pm0.12$&	$3.55\pm0.13$&	$4.47\pm0.08$\\
& Fine-Tuned    & R &$2.51\pm0.12$&$2.90\pm0.12$&$4.04\pm0.11$ \\
&            & R+ED &$3.06\pm0.14$&$3.34\pm0.13$&$4.12\pm0.11$ \\
&              & ED &$3.76\pm0.11$&$3.76\pm0.12$&$4.37\pm0.09$ \\
                           \cline{2-6}
&Multitask&ED	    &$3.63\pm0.12$&$3.83\pm0.12$&$4.49\pm0.08$\\
& EmoPrepend-1  & ED&$3.44\pm0.11$&$3.70\pm0.11$&$4.40\pm0.08$\\
&EmoPrepend-3&ED	&$3.54\pm0.11$&$3.76\pm0.11$&$4.54\pm0.07$\\
&EmoPrepend-5&ED	&$3.42\pm0.11$&$3.61\pm0.11$&$4.53\pm0.07$\\
& TopicPrepend-1&ED&$3.72\pm0.12$&$3.91\pm0.11$&$4.57\pm0.07$ \\
&TopicPrepend-3&ED	&$3.64\pm0.11$&$3.66\pm0.12$&$4.51\pm0.08$\\
&TopicPrepend-5&ED	&$3.34\pm0.12$&$3.52\pm0.12$&$4.24\pm0.09$\\
&Ensem-DM&ED	    &$3.61\pm0.11$&$3.71\pm0.12$&$4.45\pm0.08$\\
                           \cline{2-6}
&Pretrained-Large&R	&$2.94\pm0.14$&$3.12\pm0.14$&$4.23\pm0.10$\\
&                &ED&$3.47\pm0.14$&$3.56\pm0.13$&$4.41\pm0.10$\\
&Fine-Tuned-Large&ED&$3.81\pm0.12$&$3.90\pm0.12$&$4.56\pm0.08$\\

\hline
\multirow{13}{*}{Retrieval w/ BERT}
& {\it Pretrained }& R  & $3.06\pm0.13$&$3.29\pm0.13$&$4.20\pm0.10$ \\
&              & R+ED & $3.49\pm0.12$&	$3.62\pm0.12$&	$4.41\pm0.09$\\
&              & ED & $3.43\pm0.13$&	$3.49\pm0.14$&	$4.37\pm0.10$ \\
&Fine-Tuned    & R  & $2.90\pm0.13$&    $3.39\pm0.13$&  $4.36\pm0.09$\\
&            &R+ED  & $3.46\pm0.13$&    $3.90\pm0.12$&  $4.46\pm0.08$\\
&              & ED & $3.71\pm0.12$&$3.76\pm	0.12$&	$4.58\pm0.06$\\
                           \cline{2-6}
&Multitask   	&ED&$3.80\pm0.12$&$3.97\pm0.11$&$4.63\pm0.07$\\
& EmoPrepend-1  & ED &$3.93\pm0.12$&$3.96\pm0.13$&$4.54\pm0.09$\\
&EmoPrepend-3	&ED&$3.73\pm0.13$&$3.88\pm0.14$&$4.60\pm0.09$\\
&EmoPrepend-5	&ED&$4.08\pm0.10$&$4.10\pm0.11$&$4.67\pm0.07$\\
& TopicPrepend-1  & ED &$4.03\pm0.10$&$3.98\pm0.11$&$4.65\pm0.07$\\
&TopicPrepend-3	&ED&$3.73\pm0.12$&$3.84\pm0.13$&$4.52\pm0.08$\\
&TopicPrepend-5	&ED&$3.72\pm0.12$&$3.80\pm0.12$&$4.46\pm0.09$\\

\hline
\multirow{13}{*}{Generative}
& {\it Pretrained} & - &$2.31\pm0.12$&	$2.21\pm0.11$&	$3.89\pm0.12$\\
& Fine-Tuned  &  - & $3.25\pm0.12$&	$3.33\pm0.12$&	$4.30\pm0.09$ \\
                           \cline{2-6}
&Multitask &  - &$3.36\pm0.13$&$3.34\pm0.13$&$4.21\pm0.10$\\
&EmoPrepend-1  &  - &$3.16\pm0.12$&$3.19\pm0.13$&$4.36\pm0.09$\\
&EmoPrepend-3  &  - &$3.09\pm0.13$&$3.02\pm0.13$&$4.39\pm0.09$\\
&EmoPrepend-5  &  - &$3.32\pm0.12$&$3.23\pm0.12$&$4.35\pm0.09$\\
&TopicPrepend-1 & - &$3.09\pm0.13$&$3.12\pm0.13$&$4.41\pm0.08$\\
&TopicPrepend-3  &  - &$3.09\pm0.12$&$3.34\pm0.13$&$4.53\pm0.08$\\
&TopicPrepend-5  &  - &$3.46\pm0.13$&$3.68\pm0.13$&$4.60\pm0.08$\\
&Ensem-DM  &  - &$3.42\pm0.12$&$3.45\pm0.12$&$4.67\pm0.06$\\
                            \cline{2-6}
&Pretrained-Large& - &$2.84\pm0.13$&$2.97\pm0.12$&$4.01\pm0.11$\\
&Fine-Tuned-Large& - &$3.61\pm0.13$&$3.62\pm0.13$&$4.46\pm0.10$\\

\hline
{\it Gold Response} &--&--&$4.19\pm0.10$&$4.55\pm0.07$&$4.68\pm0.06$\\

\bottomrule
\end{tabular}
\caption{Human evaluation metrics from rating task for additional experiments. 
}
\label{table:human_supp_full}
\end{table*}

\begin{table*}[tb]
\begin{center}
\begin{tabular}{llllll}
\toprule
  & Model & Params, resources, train examples & Emp & Rel & Fluent \\ \midrule
\multirow{6}{*}{Retrieval} &  Pretrained-R &84.3M, 2.5 days, 8 GPUs, 1.7B&  2.8 & 3.0 & 4.1 \\
 &Pretrained-ED & same , same, +22.3k&   3.5 & 3.6 & 4.5 \\
 &\basename{} & same , + 0.5 hour, 1 GPU, +22.3k&   3.8 & 3.8 & 4.4 \\
 &Multitask &+9.6k, + 0.5 hour, 1 GPU, +22.3k&   3.6 & 3.6 & 4.5 \\
  \cline{2-6}
 &Pretrained-Large-R&86.5M, 10.5 days, 8 GPUs , 1.7B &2.9&3.1&4.2\\
 &Pretrained-Large-ED&same, same, +22.3k&3.5&3.6&4.4\\
 &\basename{}-Large&same, +1 hour, 1GPU, +22.3k&3.8&3.9&4.6\\
 \cline{2-6}
 &Pretrained-BERT-R&217M, 13.5 days, 8 GPUs , 1.7B &3.1&3.3&4.2\\
 &Pretrained-BERT-ED&same, same, +22.3k &3.4&3.5&4.4\\
 &\basename{}-BERT&same, +1 hour, 8 GPUs, +22.3k&3.7&3.8&4.6\\
 &Multitask-BERT&+9.6k, +0.5 hour, 8 GPUs, +22.3k&3.8&4.0&4.6\\

 \midrule
\multirow{4}{*}{Generative} &Pretrained &85.1M, 2 days, 32 GPUs, 1.7B&   2.3 & 2.2 & 3.9 \\
  &\basename  & same , +1 hour, 1 GPU, +22.3k&  3.3  & 3.3  & 4.3 \\
  &Multitask  &+9.6k, +1 hour, 1 GPU, +22.3k&   3.2  & 3.2  & 4.3 \\
  \cline{2-6}
  &Pretrained-Large &86.2M, 2.5 days, 32 GPUs, 1.7B&  2.8 & 3.0 & 4.0 \\
  &\basename{}-Large & same , +0.5 hour, 1 GPU, +22.3k& 3.6 & 3.6 & 4.5 \\
 \hline
\end{tabular}
\end{center}
\caption{Training resources for different models, with human 
ratings for empathy (Emp), relevance (Rel) and fluency (Fluent) for full set of experiments. Retrieval-based models use reply candidates from the ED training set (ED) or from Reddit (R). Resource comparisons are relative to the first row of each group. Fine-tuning on \datasetabbrev{} improves all scores (except for Fluency in one case) while requiring minimal additional training resources.
SEM is approximately 0.1}
\label{table:capacity_full}
\end{table*}

\subsection{Emotion Classification Results}

Our dataset can also be used to train or fine-tune an emotion classifier, as we do in our \textsc{prepend-k} and \textsc{ensem-DM+} set-ups.
To give a sense of where the difficulty falls compared to existing emotion and sentiment classification benchmarks, we reproduce the table from \citet{Deepmoji} and add results when fine-tuning the Deepmoji model on our dataset, or using a fastText classifier (Table~\ref{table:benchmarks}).

\begin{table*}[]
\begin{center}
\begin{tabular}{rrrrrrrr}
Dataset & Metric & \begin{tabular}[c]{@{}c@{}}SOTA\\(in 2017)\end{tabular} & fastText & \begin{tabular}[c]{@{}c@{}}DeepMoji\\new\end{tabular} & \begin{tabular}[c]{@{}c@{}}DeepMoji\\ full\end{tabular} & \begin{tabular}[c]{@{}c@{}}DeepMoji\\last\end{tabular} & \begin{tabular}[c]{@{}c@{}}DeepMoji\\chain-thaw\end{tabular} \\
\midrule
\textsc{SE0714} & F1 &      0.34 & 0.16 &  0.21 &	0.31 &	0.36 &	0.37 \\
\textsc{Olympic} & F1 &     0.50 & 0.38 &  0.43 &	0.50 &	0.61 &	0.61 \\
\textsc{PsychExp} & F1 &    0.45 & 0.44 &	0.32 &	0.42 &	0.56 &	0.57\\
\textsc{SS-Twitter} & Acc & 0.82 &  0.68 &	0.62 & 	0.85 &	0.87 &	0.88 \\
\textsc{SS-Youtube} & Acc & 0.86  &	0.75 &  0.75 &	0.88 &	0.92 &	0.93 \\
\textsc{SE0614} & Acc &     0.51 &     - &  0.51 &  0.54 &  0.58 &  0.58\\
\textsc{SCv1} & F1 &        0.63  &  0.60 &  0.67 &  0.65 &	0.68 &	0.69 \\
\textsc{SCv2-GEN} & F1 &    0.72 &  0.69 &  0.71 &	0.71 &	0.74 &	0.75 \\
\textsc{ED} & Acc &                 -   &   0.43 &  0.40 &  0.46 &  0.46 &  0.48 \\
\textsc{ED-cut} & Acc & - &                  0.41 & 0.36 &  0.42 &  0.44 &  0.45 \\
\bottomrule
\end{tabular}
\end{center}
\caption{Classification performance on \datasetname, with the benchmarks proposed in \cite{Deepmoji} for reference.
ED: performance on predicting the emotion label from the situation description.
ED-CUT: same, but after having removed all the situation descriptions where the target label was present.
}
\label{table:benchmarks}
\end{table*}

\end{document}